\documentclass[10pt, a4paper]{article}
\usepackage{lrec}
\usepackage{graphicx}
\usepackage{tabularx}
\usepackage{soul}

\usepackage{amssymb}

\usepackage{xcolor}

\usepackage{epstopdf}

\usepackage{hyperref}
\usepackage{xstring}

\usepackage{enumitem}
\usepackage{tikz-dependency}
\usepackage{comment}
\usepackage{lingmacros}

\makeatletter
\newcommand*{\bigcdot}{}
\DeclareRobustCommand*{\bigcdot}{%
  \mathbin{\mathpalette\bigcdot@{}}%
}
\newcommand*{\bigcdot@scalefactor}{1.25}
\newcommand*{\bigcdot@widthfactor}{1.15}
\newcommand*{\bigcdot@}[2]{%
  \sbox0{$#1\vcenter{}$}
  \sbox2{$#1\cdot\m@th$}%
  \hbox to \bigcdot@widthfactor\wd2{%
    \hfil
    \raise\ht0\hbox{%
      \scalebox{\bigcdot@scalefactor}{%
        \lower\ht0\hbox{$#1\bullet\m@th$}%
      }%
    }%
    \hfil
  }%
}
\makeatother

\definecolor{MapBlue}{HTML}{006dd2}
\definecolor{MapGreen}{HTML}{009200}
\definecolor{MapRed}{HTML}{db4955}

\title{Universal Dependencies v2: An Evergrowing Multilingual Treebank Collection}

\name{Joakim Nivre$^*$~~~Marie-Catherine de Marneffe$^\circ$~~~Filip Ginter$^\bullet$~~~Jan Haji\v{c}$^\dagger$\\[1mm]
\textbf{{\large Christopher D.~Manning$^\ddagger$~~~Sampo Pyysalo$^\bullet$~~~Sebastian Schuster$^\ddagger$}}\\[1mm]
\textbf{{\large Francis Tyers$^\diamond$~~~Daniel Zeman$^\dagger$}}\\[-1mm]}
\address{$^*$Uppsala University~~~$^\circ$The Ohio State University~~~$^\bullet$University of Turku\\[1mm]
$^\dagger$Charles University in Prague$~~~^\ddagger$Stanford University~~~$^\diamond$Indiana University\\[2mm]
$^*$joakim.nivre@lingfil.uu.se~~~$^\circ$demarneffe.1@osu.edu~~~$^\bullet$\{figint,sampo.pyysalo\}@utu.fi\\[1mm]
$^\dagger$\{jan.hajic,daniel.zeman\}@mff.cuni.cz~~~$^\ddagger$\{manning,sebschu\}@stanford.edu~~~$^\diamond$ftyers@iu.edu\\
}

\abstract{
Universal Dependencies is an open community effort to create cross-linguistically consistent treebank annotation for many languages within a dependency-based lexicalist framework. The annotation consists in a linguistically motivated word segmentation; a morphological layer comprising lemmas, universal part-of-speech tags, and standardized morphological features; and a syntactic layer focusing on syntactic relations between predicates, arguments and modifiers. In this paper, we describe version 2 of the guidelines (UD v2), discuss the major changes from UD v1 to UD v2, and give an overview of the currently available treebanks for 90 languages. \\ \newline  }

\begin{document}

\maketitleabstract

\section{Introduction}

Universal Dependencies (UD) is a project that is developing cross-linguistically consistent treebank annotation for many languages, with the goal of facilitating multilingual parser development and research on parsing and cross-lingual learning. The annotation scheme is based on an evolution of (universal) Stanford dependencies \cite{demarneffe06,demarneffe08,demarneffe14}, Google universal part-of-speech tags \cite{petrov12lrec}, and the Interset interlingua for morphosyntactic tagsets \cite{zeman08}. The general philosophy is to provide a universal inventory of categories and guidelines to facilitate consistent annotation of similar constructions across languages, while allowing language-specific extensions when necessary.

\begin{figure}[b]
    \centering
    \includegraphics[width=0.44\textwidth]{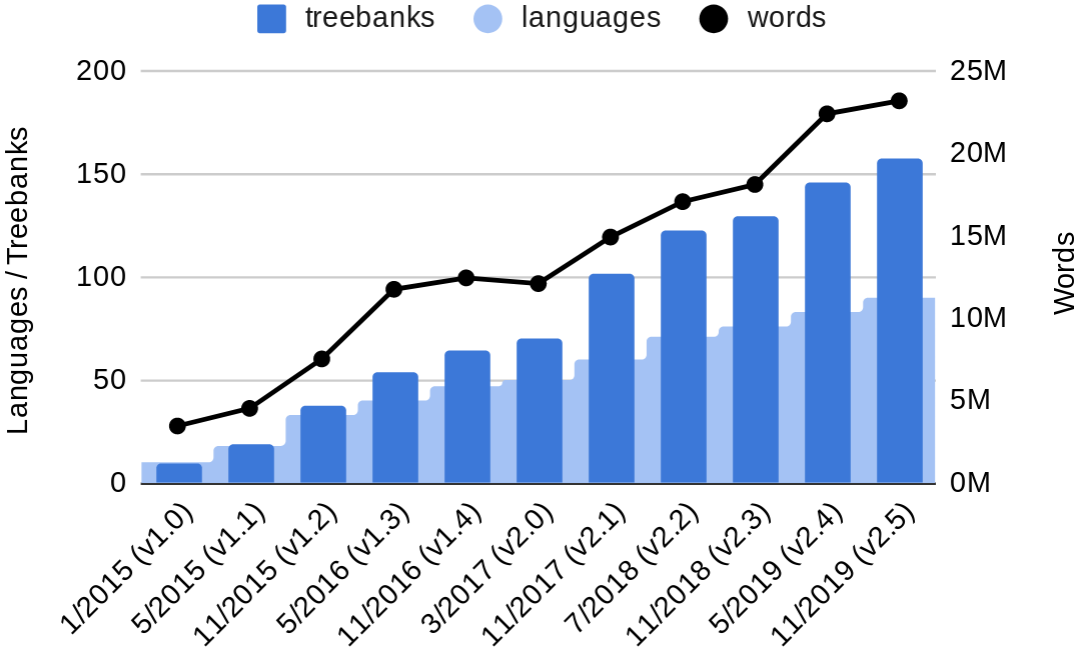}
    \caption{Number of languages, treebanks and words in UD from v1.0 to v2.5.}
    \label{fig:growth}
\end{figure}

\begin{figure*}[t]
\centering
\includegraphics[width=\textwidth]{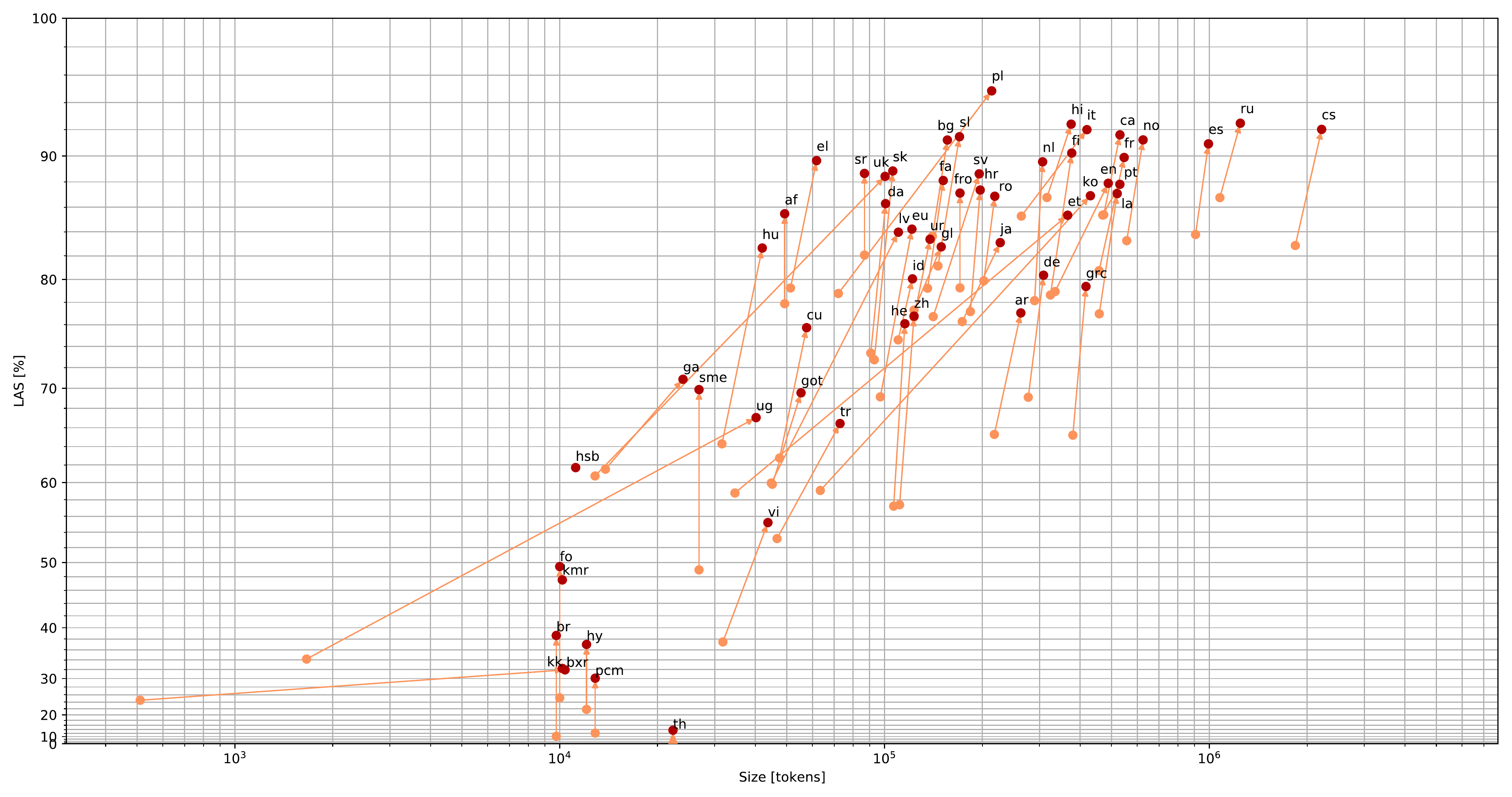}
\vspace{-10mm}
\caption{Increase in available data (x-axis) and labeled attachment score (y-axis) from the baseline of the CoNLL 2017 shared task (orange) to the best result of the CoNLL 2018 shared task (red);
pairs labeled by ISO language codes.}
\label{fig:parsing}
\end{figure*}

The project started in 2014 and has developed into an open community effort with a very rapid growth, both in terms of the number of researchers contributing to the project, which now exceeds 300, and in terms of the number of languages represented by treebanks, which is  approaching 100. An early snapshot of this development can be found in \newcite{nivre16lrec}, which describes version 1 of the UD guidelines (UD v1) and the treebank resources available in UD v1.2. Since then, there has been one major change of the guidelines, from UD v1 to UD v2, and the number of treebanks has more than quadrupled. Figure~\ref{fig:growth} shows the growth in number of languages, treebanks and annotated words from UD v1.0 to UD v2.5. During the same period, the number of downloads or accesses at the official repository at \url{https://lindat.cz} has grown to 46439.\footnote{November 25, 2019.} The UD resources have also made a significant impact on NLP research, most notably for multilingual dependency parsing through two editions of CoNLL shared tasks \cite{zeman17,zeman18conll}, which have created a new generation of parsers that handle a large number of languages and that parse from raw text rather than relying on pre-tokenized input. Figure~\ref{fig:parsing} visualizes the increase in available data resources and parsing scores for all languages involved in both tasks. 

This paper provides an up-to-date description of the project, focusing on the annotation guidelines, especially on the major changes from UD v1 to v2, and on the existing treebank resources. For more information on the project motivation and history, we refer to \newcite{nivre16lrec}. For more information about UD treebanks and applications of these resources, we refer to the proceedings of the UD workshops held annually since 2017 \cite{udw17,udw18,udw19}.

\section{Annotation Scheme}
In this section, we give a brief introduction to the UD annotation scheme.  
For more details, we refer to the documentation on the UD website.\footnote{https://universaldependencies.org/guidelines.html}

\subsection{Tokenization and Word Segmentation}
\label{sec:word}
UD is based on a lexicalist view of syntax, which means that dependency relations hold between words,
and that morphological features are encoded as properties of words with no attempt at segmenting words into morphemes. However, it is important to note that the basic units of annotation are syntactic words (not phonological or orthographic words), which means that it is often necessary to split off clitics, as in Spanish \emph{d\'{a}melo} = \emph{da me lo}, and undo contractions, as in French \emph{au} = \emph{\`{a} le}. We refer to such cases as \emph{multiword tokens} because a single orthographic token corresponds to multiple (syntactic) words. In exceptional cases, it may be necessary to go in the other direction, and combine several orthographic tokens into a single syntactic word (see Section~\ref{sec:word2}).

%

\subsection{Morphological Annotation}
\label{sec:morphology}
The morphological specification of a (syntactic) word in the UD scheme consists of three levels of representation:
\begin{enumerate}[topsep=3pt,noitemsep]
\item A lemma representing the base form of the word.
\item A part-of-speech tag representing the grammatical category of the word.
\item A set of features representing lexical and grammatical properties associated with the particular word form.
\end{enumerate}
The lemma is the canonical form of the word, which is the form typically found in dictionaries. In agglutinative languages, this is typically the form with no inflectional affixes; in fusional languages, the lemma is usually the result of a language-particular convention. 
The list of universal part-of-speech tags is a fixed list containing 17 tags, shown in Table~\ref{tab:label}. Languages are not required to use all tags, but the list cannot be extended to cover language-specific categories. Instead, more fine-grained classification of words can be achieved via the use of features, which specify
additional information about morphosyntactic properties. 
We provide an inventory of features that are attested in multiple languages and need to be encoded in a uniform way, listed in Table~\ref{tab:label}. Users can extend this set of universal features and add language-specific features when necessary.

\begin{table*}[tb]
\centering
\begin{tabular}{|l|ll|lll|}
\hline
&&& \multicolumn{3}{c|}{\textbf{Syntactic Relations}} \\
\cline{4-6}
& \multicolumn{2}{c|}{\textbf{Features}} & \multicolumn{2}{c|}{\textbf{Clausal}} & \\ 
\textbf{PoS Tags} & \textbf{Inflectional} & \textbf{Lexical} & \textbf{Core} & \multicolumn{1}{l|}{\textbf{Non-Core}} & \textbf{Nominal} \\
\hline
ADJ & Animacy & Abbr & nsubj & advcl & acl \\
ADP & Aspect & Foreign & csubj & advmod & amod \\
ADV & Case & NumType & ccomp & aux & appos \\
AUX & Clusivity & Poss & iobj & cop & case \\
CCONJ & Definite & PronType & obj & discourse & clf \\
DET & Degree & Reflex & xcomp & dislocated & det \\
INTJ & Evident &Typo & & expl & nmod \\
NOUN & Gender & & & mark & nummod \\
NUM & Mood & & & obl & \\
PART & NounClass & & & vocative & \\
\cline{4-6}
PRON & Number & & \textbf{Linking} & \textbf{MWE} & \textbf{Special} \\
\cline{4-6}
PROPN & Person & & cc & compound & dep \\
PUNCT & Polarity & & conj & fixed & goeswith \\
SCONJ & Polite & & list & flat & orphan \\
SYM & Tense & & parataxis & & punct \\
VERB & VerbForm & & & & reparandum \\
X & Voice & & & & root \\
\hline
\end{tabular}
\caption{Universal part-of-speech tags (left), morphological features (middle) and syntactic relations (right).}
\label{tab:label}
\end{table*}


\subsection{Syntactic Annotation}
\label{sec:syntax}
Syntactic annotation in the UD scheme consists of typed dependency relations between words. The \emph{basic} syntactic representation forms a tree rooted in one word, normally the main clause predicate, on which all other words of the sentence are dependent. In addition to the basic representation, which is obligatory for all UD treebanks, it is possible to give an \emph{enhanced} dependency representation, which adds (and in a few cases changes) relations in order to give a more complete basis for semantic interpretation. We will focus here on the basic representation and return to the enhanced representation when discussing changes in UD v2.

\begin{figure}[tb]
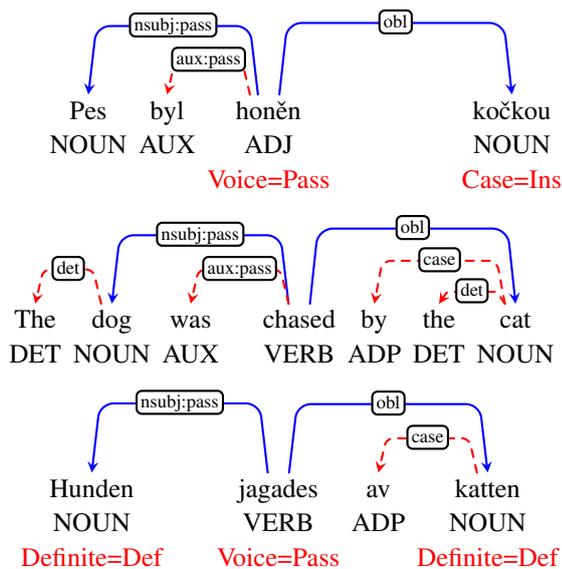

\hspace{-3mm}
\scalebox{1}{
\begin{tabular}{c}
\begin{dependency}
\begin{deptext}
\&[18pt] Pes \& byl \&hon\v{e}n \&[45pt] ko\v{c}kou \\[0.1cm]
\& NOUN \&[0pt] AUX \& ADJ \& NOUN \\[0.1cm]
\& \&  \& \textcolor{red}{Voice=Pass}\& \textcolor{red}{Case=Ins} \& \\
\end{deptext}
\depedge[edge style=blue, thick]{4}{2}{nsubj:pass}
\depedge[edge style=red, thick, dashed, edge start x offset=-3pt]{4}{3}{aux:pass}
\depedge[edge style=blue, thick, edge unit distance=6ex]{4}{5}{obl}
\end{dependency}
\\
\begin{dependency}
\begin{deptext}
The \& dog \& was \&[12pt] chased \& by \& the \& cat \\[0.1cm]
DET \& NOUN \& AUX \& VERB \& ADP \& DET \& NOUN \\
\end{deptext}
\depedge[edge style=red, thick, dashed]{2}{1}{det}
\depedge[edge style=blue, thick]{4}{2}{nsubj:pass}
\depedge[edge style=red, thick, dashed]{4}{3}{aux:pass}
\depedge[edge style=red, thick, dashed, edge unit distance=1.8ex]{7}{5}{case}
\depedge[edge style=red, thick, dashed, edge unit distance=1ex]{7}{6}{det}
\depedge[edge style=blue, thick, edge unit distance=2.1ex]{4}{7}{obl}
\end{dependency}
\\
\begin{dependency}
\begin{deptext}
\&[6pt] Hunden \&[16pt] jagades \&[0pt] av \& katten \& \\[0.1cm]
\& NOUN \& VERB \&[0pt] ADP \& NOUN \& \\[0.1cm]
\& \textcolor{red}{Definite=Def}  \& \textcolor{red}{Voice=Pass} \&[0pt] \& \textcolor{red}{Definite=Def} \& \\
\end{deptext}
\depedge[edge style=blue, thick, edge unit distance=5.8ex]{3}{2}{nsubj:pass}
\depedge[edge style=red, thick, dashed]{5}{4}{case}
\depedge[edge style=blue, thick]{3}{5}{obl}
\end{dependency}
\end{tabular}
}
\caption{Parallel sentences in Czech, English and Swedish. Common syntactic relations in blue, differences in morphosyntactic encoding highlighted in red. 
The Czech passive participle has both adjectival and verbal features; it is tagged ADJ due to its similarity to adjectives.}
\label{fig:passive}
\end{figure}

The syntactic analysis in UD gives priority to predicate-argument and modifier relations that hold directly between content words, as opposed to being mediated by function words. The rationale is that this makes more transparent what grammatical relations are shared across languages, even when the languages  differ in the way that they use word order, function words or morphological inflection to encode these relations. This is illustrated in Figure~\ref{fig:passive}, which shows three parallel sentences in Czech, English and Swedish. In all three cases, there is a passive predicate with a subject and an oblique modifier (the relations marked in solid blue), but the languages differ in how they encode certain grammatical categories (marked in dashed red): definiteness is indicated by a separate function word (the article \emph{the}) in English, by a morphological inflection in Swedish and not at all in Czech; passive is expressed by a periphrastic construction involving an auxiliary and a participle in English, by a morphological inflection in Swedish, and by a combination of these strategies in Czech (because the participle is unique to the passive construction); and the oblique modifier is introduced by a preposition in English and Swedish but marked by instrumental case in Czech.

UD provides a taxonomy of 37 universal relation types to classify syntactic relations, as shown in Table~\ref{tab:label}. The taxo\-nomy distinguishes between relations that occur at the clause level (linked to a predicate) and those that occur in noun phrases (linked to a nominal head). At the clause level, a distinction is made between core arguments (essentially subjects and objects) and all other dependents \cite{thompson97,andrews07}. It is important to note that not all relations in the taxonomy are syntactic dependency relations in the narrow sense. First, there are special relations for function words like determiners, classifiers, adpositions, auxiliaries, copulas and subordinators, whose dependency status is controversial. In addition, there are a number of special relations for linking relations (including coordination), certain types of multiword expressions, and special phenomena like ellipsis, disfluencies, punctuation and typographical errors. Many of these relations cannot plausibly be interpreted as syntactic head-dependent relations, and should rather be thought of as technical devices for encoding flat structures in the form of a tree.

The inventory of universal relation types is fixed, but subtypes can be added in individual languages to capture additional distinctions that are useful. This is illustrated in Figure~\ref{fig:passive}, where the relations \textsc{nsubj}%
\footnote{Syntactic relations in UD are normally written in all lowercase, as shown in Table~\ref{tab:label}, but in this paper we use small capitals in running text for clarity.}
(nominal subject) and \textsc{aux} (auxiliary) are subtyped to \textsc{nsubj:pass} and \textsc{aux:pass} to capture properties of passive constructions.



\section{Changes from UD v1 to UD v2}
\label{sec:v2}

We now discuss the most important changes from UD v1 to UD v2. 
More information about these changes can be found on the UD website.\footnote{https://universaldependencies.org/v2/summary.html}

\subsection{Tokenization and Word Segmentation}
\label{sec:word2}

In UD v1, word-internal spaces were not allowed. This restriction has now been lifted in two circumstances:
\begin{enumerate}[topsep=3pt,noitemsep]
\item For languages with writing systems that use spaces to mark units smaller than words (typically syllables), spaces are allowed in any word; the phenomenon has to be declared in the language-specific documentation.
\item For other languages, spaces are allowed only for a restricted list of exceptions like numbers (\emph{100 000}) and abbreviations (\emph{i.~e.}); the latter have to be listed explicitly in the language-specific documentation.
\end{enumerate}
The first case was deemed necessary, because in languages like Vietnamese all polysyllabic words would otherwise have to be annotated as fixed multiword expressions, which would seriously distort the syntactic representations compared to other languages. The second case is more a matter of convenience, but it seemed useful to allow \emph{multitoken words} -- a single (syntactic word) corresponding to multiple orthographic tokens -- as well as multiword tokens, 
although this option should be used very restrictively.

\begin{table}[t]
\centering
\begin{footnotesize}
\renewcommand{\tabcolsep}{2pt}
\begin{tabular}{|l|l||l|l|}
\hline
\multicolumn{2}{|l||}{\textbf{Feature}} & \multicolumn{2}{l|}{\textbf{Value(s)}} \\
\hline
\textbf{Old} & \textbf{New} & \textbf{Old} & \textbf{New} \\
\hline
& Clusivity && Ex, In \\
& Evident   && Nfh \\
& NounClass && Bantu1--23, Wol1--12, \dots{} \\
& Polite    && Infm, Form, Elev, Humb \\
& Abbr      && Yes \\
& Foreign   && Yes \\
& Typo      && Yes \\
\hline
Animacy &&& Hum \\
Case &&& Equ, Cmp, Cns, Per \\
Degree &&& Equ \\
Definite &&& Spec \\
Number &&& Count, Tri, Pauc, Grpa, Grpl, Inv \\
VerbForm &&& Gdv, Vnoun \\
Mood &&& Prp, Adm \\
Aspect &&& Iter, Hab \\
Voice &&& Mid, Antip, Dir, Inv \\
PronType &&& Emp, Exc \\
Person &&& 0, 4 \\
\hline
Negative & Polarity &&\\
\hline
Aspect && Pro & Prosp \\
VerbForm && Trans & Conv \\
Definite && Red & Cons \\
\hline
\end{tabular}
\end{footnotesize}
\caption{Revisions to morphological features and values in UD v2: new features (group 1), new values (group 2), and renamed features and values (groups 3 and 4).}
\label{tab:new-features}
\end{table}

\subsection{Morphological Annotation}
\label{sec:morphology2}
The universal part-of-speech tagset is essentially the same in UD v2 as in UD v1, but the tag for coordinating conjunctions has been renamed from CONJ to CCONJ\footnote{The motivation is to make it parallel to SCONJ (for subordinating conjunctions), more similar to the syntactic relation \textsc{cc} with which it often cooccurs, and less similar to the relation \textsc{conj} with which it practically never cooccurs.} and the guidelines have been modified slightly for three tags:
\begin{enumerate}[topsep=3pt,noitemsep]
\item The use of AUX is extended from auxiliary verbs in a narrow sense to also include copula verbs and nonverbal TAME particles (tense, aspect, mood, evidentiality, and, sometimes, voice or polarity particles).
\item The use of PART is limited to a small set of words that must be listed in the language-specific documentation.
\item The distinction between PRON and DET is made more flexible to accommodate cross-linguistic variation.
\end{enumerate}
The inventory of universal morphological features has been extended with new features and new values for existing features. In addition, a few features and feature values have been renamed or removed. These changes, which are summarized in Table~\ref{tab:new-features}, are motivated by the addition of new languages to UD as well as an effort to harmonize UD with the UniMorph project \cite{sylak15}.



\subsection{Syntactic Annotation}
\label{sec:syntax2}
Although most syntactic relations are the same in UD v2 as in UD v1, the guidelines have often been improved by providing more explicit criteria and examples from multiple languages. Here we only list cases where relations have been removed, added or renamed, or where the use of an existing relation has changed significantly.

\paragraph{Clauses and Dependents of Predicates}
As explained earlier, UD assumes a distinction between core and non-core dependents of predicates. For nominal core arguments, UD v1 used the labels \textsc{nsubj}, \textsc{dobj} and \textsc{iobj}. These relations remain conceptually unchanged, but the second label has been changed from \textsc{dobj} to \textsc{obj}, because this seems to better convey the intended interpretation of ``second core argument" or ``P/O argument" (without connection to specific cases or semantic roles). In addition, the \textsc{nsubjpass} label for passive subjects is removed, and passive subjects are subsumed under the \textsc{nsubj} relation, but with a strong recommendation to use the subtype \textsc{nsubj:pass} for languages where the distinction is relevant. Analogously, the relations \textsc{csubjpass} (for clausal passive subject) and \textsc{auxpass} (for passive auxiliary) are now subsumed under \textsc{csubj} and \textsc{aux} (with possible subtypes \textsc{csubj:pass} and \textsc{aux:pass}).

The second change in this area concerns the analysis of \emph{oblique} nominals at the clause level, that is, nominal expressions that are dependents of predicates but not core arguments, and which are typically accompanied by case marking in the form of adpositions or oblique morphological case. In UD v1, such expressions were subsumed under the \textsc{nmod} relation (for nominal modifier), which also applies to nominal expressions that modify other nominals and are not dependents of predicates at the clause level. This violated a fundamental principle of UD, namely that distinct labels should be used for dependents of nominals and dependents of predicates, even if the overt form of the modifier is the same. In UD v2, the \textsc{obl} relation is therefore used for oblique nominals at the clause level, while the \textsc{nmod} relation is reserved for nominals modifying other nominal expressions. The distinction is illustrated in (\ref{clause}) and (\ref{nominal}), which also show that the core/non-core distinction is only applied at the clause level. Hence, both the \textsc{nsubj} and the \textsc{obl} relations in the clause example correspond to \textsc{nmod} relations in the nominal example.

\enumsentence{
\raisebox{-2.5mm}{
\begin{dependency}
\begin{deptext}
she \& suddenly \& went \& to \& Paris \\[0.1cm]
PRON \& ADV \& VERB \& ADP \& PROPN \\
\end{deptext}
\depedge[edge style=thick]{3}{1}{nsubj}
\depedge[edge style=thick]{3}{2}{advmod}
\depedge[edge style=thick]{5}{4}{case}
\depedge[edge style=thick]{3}{5}{obl}
\end{dependency}
 }
 \label{clause}
 }

\enumsentence{
\raisebox{-2.5mm}{
\begin{dependency}
\begin{deptext}
her \& sudden \& trip \& to \& Paris \\[0.1cm]
PRON \& ADJ \& NOUN \& ADP \& PROPN \\
\end{deptext}
\depedge[edge style=thick]{3}{1}{nmod}
\depedge[edge style=thick]{3}{2}{amod}
\depedge[edge style=thick]{5}{4}{case}
\depedge[edge style=thick]{3}{5}{nmod}
\end{dependency}
 }
 \label{nominal}
 }

The final modification in the annotation of clause structure is a more restricted application of the \textsc{cop} relation. In UD v2, the \textsc{cop} relation is restricted to function words (verbal or nonverbal) whose sole function is to link a nonverbal predicate to its subject and which does not add any meaning other than grammaticalized TAME categories. The range of constructions that are analyzed using the \textsc{cop} relation is subject to language-specific variation but can be identified using universal criteria described in the guidelines.

\paragraph{Coordination}
The question of whether and how coordination can be analyzed as a dependency structure is a vexed one \cite{popel13,gerdes15}. UD treats coordination as an essentially symmetric relation, and uses the special \textsc{conj} relation to connect all non-first conjuncts to the first one. In this respect, UD v2 is exactly the same as UD v1, but UD v2 differs by attaching coordinating conjunctions (\textsc{cc}) and punctuation (\textsc{punct}) inside coordinated structures to the immediately succeeding conjunct (instead of the first conjunct as in UD v1), following the approach of \newcite{ross67}, as illustrated in (\ref{coord}).

\enumsentence{
\raisebox{-2.5mm}{
\begin{dependency}
\begin{deptext}
bacon \& , \& lettuce \& and \& tomato \\[0.1cm]
NOUN \& PUNCT \& NOUN \& CCONJ \& NOUN \\
\end{deptext}
\depedge[edge style=thick]{3}{2}{punct}
\depedge[edge style=thick]{1}{3}{conj}
\depedge[edge style=thick]{5}{4}{cc}
\depedge[edge style=thick, edge unit distance=1.0em]{1}{5}{conj}
\end{dependency}
}
\label{coord}
}

\paragraph{Ellipsis}
The analysis of elliptical constructions like gapping is completely different in UD v2 compared to UD v1. Let us first note that most cases of ellipsis are simply treated by ``promoting'' a dependent of the elided element to take its place in the syntactic structure. Thus, adjectival modifiers or even determiners can head nominals if the head noun is omitted. Similarly, auxiliary verbs can head clauses in constructions like VP ellipsis. However, in cases like gapping, this yields a rather unsatisfactory analysis where one core argument is typically attached to another. UD v2 therefore uses a special relation \textsc{orphan} to indicate that this is an anomalous structure where the dependent is really a sibling of the word to which is it attached. As illustrated in (\ref{ellipsis}), this gives an underspecified analysis of the predicate-argument structure, which can be fully resolved in the enhanced representation (see Section~\ref{sec:enhanced}).

\enumsentence{
\raisebox{-2.5mm}{
\begin{dependency}
\begin{deptext}
she \& drank \& coffee \& and \& he \& tea \\[0.1cm]
PRON \& VERB \& NOUN \& CCONJ \& PRON \& NOUN \\
\end{deptext}
\depedge[edge style=thick]{2}{1}{nsubj}
\depedge[edge style=thick]{2}{3}{obj}
\depedge[edge style=thick]{5}{4}{cc}
\depedge[edge style=thick,edge unit distance=2ex]{2}{5}{conj}
\depedge[edge style=thick]{5}{6}{orphan}
\end{dependency}
}
\label{ellipsis}
}

The choice of which dependent to promote is determined by an obliqueness hierarchy (where subjects precede objects) described in the guidelines. This new analysis of gapping is  superior to the UD v1 analysis (which used a \textsc{remnant} relation), because it preserves the integrity of the two clauses and introduces fewer non-projective dependencies.

\paragraph{Functional Relations}
UD v2 also includes some changes in the annotation of functional relations, that is, relations holding between a function word or grammatical marker and its host (mostly a verb or noun). More specifically:
\begin{enumerate}[topsep=3pt,noitemsep]
\item A new relation \textsc{clf} is added for nominal classifiers.
\item The \textsc{aux} relation is extended from auxiliary verbs in a narrow sense to also include nonverbal TAME particles in analogy with the extended use of the part-of-speech tag AUX (see Section~\ref{sec:morphology2}).
\item The \textsc{auxpass} relation is subsumed under the \textsc{aux} relation (see above).
\item The \textsc{cop} relation is restricted to pure linking words (see above).
\item The \textsc{neg} relation is removed from the set of universal relations, and polarity is instead encoded in a feature (see Section~\ref{sec:morphology2}).
\end{enumerate}

\subsubsection{Multiword Expressions}

The guidelines for annotation of multiword expressions have been thoroughly revised in UD v2. Multiword expressions that are morphosyntactically regular (and only exhibit semantic non-compositionality) normally do not receive any special treatment at all. Hence, the UD guidelines in this area only apply to a few subtypes of the many phenomena that have been discussed in the literature on multiword expressions.

The first subtype is compounding. The relation \textsc{compound} is used for any kind of lexical compounds: noun compounds such as \emph{phone book}, but also verb and adjective compounds, such as the serial verbs that occur in many languages,
%
%
or a Japanese light verb construction such as \emph{benky\={o} suru} (``to study''). The compound relation is also used for phrasal verbs, such as  \emph{put up}: \textsc{compound}(\emph{put,~up}). Despite operating at the lexical level, compounds are regular headed constructions, as illustrated in (\ref{compound}).
%
This behavior distinguishes compounds from the other two types of multiword expressions.

\enumsentence{
\raisebox{-2.5mm}{
\begin{dependency}
   \begin{deptext}[column sep=0.6cm]
     hate \& speech \& detection \\
     NOUN \& NOUN \& NOUN \\
   \end{deptext}
   \depedge[edge style=thick]{3}{2}{compound}
   \depedge[edge style=thick]{2}{1}{compound}
 \end{dependency}
}
\label{compound}
}

The second subtype is fixed expressions, highly grammaticalized 
expressions that typically behave as function words or short adverbials, for which the relation \textsc{fixed} is used. The name and rough scope of usage is borrowed from the fixed expressions category of \newcite{sag02}.%
\footnote{This relation was called \textsc{mwe} in UD v1, but the name was found to be misleading as the relation only applies to a very small subset
of multiword expressions.}
Fixed multiword expressions are annotated with a flat structure. Since there is no clear basis for internal syntactic structure, we adopt the convention of always attaching subsequent words to the first one with the \textsc{fixed} label, as shown in (\ref{fixed}).
%
%

\enumsentence{
\raisebox{-2.5mm}{
\begin{dependency}
   \begin{deptext}[column sep=0.6cm]
dogs \& as \& well \& as \& cats \\
NOUN \& ADP \& ADV \& ADP \& NOUN \\
   \end{deptext}
   \depedge[edge style=thick]{2}{4}{fixed}
   \depedge[edge style=thick]{2}{3}{fixed}
   \depedge[edge style=thick]{5}{2}{cc}
   \depedge[edge style=thick]{1}{5}{conj}
 \end{dependency}
}
\label{fixed}
}

As with other clines of grammaticalization, it is not always clear where to draw the line between giving a regular syntactic analysis versus a fixed expression analysis of a conventionalized expression. In practice, the best solution is to be conservative and to prefer a regular syntactic analysis except when an expression is highly opaque and clearly does not have internal syntactic structure (except from a historical perspective).


The final subtype is headless multiword expressions analyzed with the relation \textsc{flat}. This class is less clearly recognized in most grammars of human languages, but in practice there are many linguistic constructions with a sequence of words that do not have any clear synchronic grammatical structure but are not fixed expressions. These include names, dates, and calqued expressions from other languages. We again adopt the convention that in these cases subsequent words are attached to the first word with the \textsc{flat} relation, as exemplified in (\ref{flat}).

\enumsentence{
\raisebox{-2.5mm}{
\begin{dependency}
   \begin{deptext}[column sep=0.4cm]
     Hillary \& Rodham \& Clinton 
\\
PROPN \& PROPN \& PROPN \\
   \end{deptext}
   \depedge[edge style=thick]{1}{2}{flat}
   \depedge[edge style=thick]{1}{3}{flat}
 \end{dependency}
}
\label{flat}
}

This relation replaces two more specific relations from UD v1, \textsc{name} and \textsc{foreign}. Subtypes like \textsc{flat:name} and \textsc{flat:foreign} can be used in cases where a flat analysis is appropriate for complex names and foreign expressions.

\begin{table*}[t]
    \centering
\renewcommand{\tabcolsep}{3pt}
\scalebox{0.85}{
    \begin{tabular}{|lrrr|lrrr|lrrr|}
    \hline
    \textbf{Language}  & \textbf{\#} & \textbf{Sents} & \textbf{Words} & \textbf{Language}  & \textbf{\#} & \textbf{Sents} & \textbf{Words} & \textbf{Language}  & \textbf{\#} & \textbf{Sents} & \textbf{Words} \\
    \hline
Afrikaans &1 &1,934 &49,276 &German &4 &208,440 &3,753,947 &Old Russian &2 &17,548 &168,522 \\
Akkadian &1 &101 &1,852 &Gothic &1 &5,401 &55,336 &Persian &1 &5,997 &152,920 \\
Amharic &1 &1,074 &10,010 &Greek &1 &2,521 &63,441 &Polish &3 &40,398 &499,392 \\
Ancient Greek &2 &30,999 &416,988 &Hebrew &1 &6,216 &161,417 &Portuguese &3 &22,443 &570,543 \\
Arabic &3 &28,402 &1,042,024 &Hindi &2 &17,647 &375,533 &Romanian &3 &25,858 &551,932 \\
Armenian &1 &2502 &52630 &Hindi English &1 &1,898 &26,909 &Russian &4 &71,183 &1,262,206 \\
Assyrian &1 &57 &453 &Hungarian &1 &1,800 &42,032 &Sanskrit &1 &230 &1,843 \\
Bambara &1 &1,026 &13,823 &Indonesian &2 &6,593 &141,823 &Scottish Gaelic &1 &2,193 &42,848 \\
Basque &1 &8,993 &121,443 &Irish &1 &1,763 &40,572 &Serbian &1 &4,384 &97,673 \\
Belarusian &1 &637 &13,325 &Italian &6 &35,481 &811,522 &Skolt S\'ami &1 &36 &321 \\
Bhojpuri &1 &254 &4,881 &Japanese &4 &67,117 &1,498,560 &Slovak &1 &10,604 &106,043 \\
Breton &1 &888 &10,054 &Karelian &1 &228 &3,094 &Slovenian &2 &11,188 &170,158 \\
Bulgarian &1 &11,138 &156,149 &Kazakh &1 &1,078 &10,536 &Spanish &3 &34,693 &1,004,443 \\
Buryat &1 &927 &10,185 &Komi Permyak &1 &49 &399 &Swedish &3 &12,269 &206,855 \\
Cantonese &1 &1,004 &13,918 &Komi Zyrian &2 &327 &3,463 &Swedish Sign Language &1 &203 &1,610 \\
Catalan &1 &16,678 &531,971 &Korean &3 &34,702 &446,996 &Swiss German &1 &100 &1,444 \\
Chinese &5 &12,449 &285,127 &Kurmanji &1 &754 &1,0260 &Tagalog &1 &55 &292 \\
Classical Chinese &1 &15,115 &74,770 &Latin &3 &41,695 &582,336 &Tamil &1 &600 &9,581 \\
Coptic &1 &1,575 &40,034 &Latvian &1 &13,643 &219,955 &Telugu &1 &1,328 &6,465 \\
Croatian &1 &9,010 &199,409 &Lithuanian &2 &3,905 &75,403 &Thai &1 &1,000 &22,322 \\
Czech &5 &127,507 &2,222,163 &Livvi &1 &125 &1,632 &Turkish &3 &9,437 &91,626 \\
Danish &1 &5,512 &100,733 &Maltese &1 &2,074 &44,162 &Ukrainian &1 &7,060 &122,091 \\
Dutch &2 &20,916 &306,503 &Marathi &1 &466 &3,849 &Upper Sorbian &1 &646 &11,196 \\
English &7 &35,791 &620,509 &Mby\'a Guaran\'i &2 &1,144 &13,089 &Urdu &1 &5,130 &138,077 \\
Erzya &1 &1,550 &15,790 &Moksha &1 &65 &561 &Uyghur &1 &3,456 &40,236 \\
Estonian &2 &32,634 &465,015 &Naija &1 &948 &12,863 &Vietnamese &1 &3,000 &43,754 \\
Faroese &1 &1,208 &10,002 &North S\'ami &1 &3,122 &26,845 &Warlpiri &1 &55 &314 \\
Finnish &3 &34,859 &377,619 &Norwegian &3 &42,869 &666,984 &Welsh &1 &956 &16,989 \\
French &7 &45,074 &1,157,171 &Old Church Slavonic &1 &6,338 &57,563 &Wolof &1 &2,107 &44,258 \\
Galician &2 &4,993 &164,385 &Old French &1 &17,678 &170,741 &Yoruba &1 &100 &2,664 \\
\hline
    \end{tabular}
}
    \caption{Languages in UD v2.5 with number of treebanks (\#), sentences (Sents) and words (Words).}
    \label{tab:treebanks}
\end{table*}

\subsection{Enhanced Dependencies}
\label{sec:enhanced}

UD v2 now also provides guidelines for \emph{enhanced} dependency graphs.
With a few exceptions, enhanced graphs consist of all the syntactic relations in the \emph{basic} dependency
tree and may contain additional relations and nodes that make otherwise implicit relations between tokens explicit,
with the purpose of facilitating downstream natural language understanding tasks.
The guidelines are based on the \textit{CCprocessed} Stanford dependencies representation \cite{demarneffe06} and a proposal for \textit{enhanced} dependencies \cite{schuster16}, and define five types of enhancements. For more information, we refer to the documentation on the UD website.\footnote{https://universaldependencies.org/u/overview/enhanced-syntax.html}

\paragraph{Null Nodes for Elided Predicates}
For sentences with elided predicates, in the {basic} representation,
one word is promoted to be the head of the clause and all words that would have been a sibling of
the promoted word if no predicate had been elided are attached with the \textsc{orphan} relation (see Section~\ref{sec:syntax2}).
The {enhanced} representation for sentences with gapping contains additional null nodes representing elided predicates. Arguments
and modifiers of the elided predicate are attached to the null nodes, as illustrated in (\ref{gapping}), which
contains a null node (\textbf{E5.1}) and relations between the null node and the arguments in the second clause.

\enumsentence{
\raisebox{-2.5mm}{
\scalebox{0.9}{
\begin{dependency}
\begin{deptext}
she \& drank \& coffee \& and \& he \& {\bf E5.1} \& tea \\
PRON \& VERB \& NOUN \& CCONJ \& PRON \& VERB \& NOUN \\
\end{deptext}
\depedge[edge style=thick]{2}{1}{nsubj}
\depedge[edge style=thick]{2}{3}{obj}
\depedge[edge style=thick]{6}{4}{cc}
\depedge[edge style=thick]{6}{5}{nsubj}
\depedge[edge style=thick, edge unit distance=1em]{2}{6}{conj}
\depedge[edge style=thick]{6}{7}{obj}
\end{dependency}
}
}
\label{gapping}
}

\paragraph{Propagation of Conjuncts}

Conjoined predicates often share dependents (e.g., a subject) and conjoined dependents share a head.
In (\ref{prop}), the two predicates (\emph{buys} and \emph{sells}) share the subject (\emph{the store}) and object (\emph{cameras}).
The shared status of dependents and governors is made explicit in the {enhanced} representation through additional relations, such as the \textsc{nsubj} and \textsc{obj} relations
below the sentence.\footnote{The placement of arcs above and below the sentence, respectively, is only for perspicuity and does not imply any difference in status between different types of arcs.}

\enumsentence{
\raisebox{-8mm}{
\begin{dependency}
\begin{deptext}
the \& store \& buys \& and \& sells \& cameras \\
ADP \& NOUN \& VERB \& CCONJ \& VERB \& NOUN \\
\end{deptext}
\depedge[edge style=thick]{2}{1}{det}
\depedge[edge style=thick]{3}{2}{nsubj}
\depedge[edge style=thick]{5}{4}{cc}
\depedge[edge style=thick]{3}{5}{conj}
\depedge[edge style=thick]{3}{6}{obj}
\depedge[edge style=thick, edge below]{5}{6}{obj}
\depedge[edge style=thick, edge below, edge unit distance=0.45em]{5}{2}{nsubj}

\end{dependency}
}
\label{prop}
}

\paragraph{Controlled and Raised Subjects}
For sentences with control or raising predicates, in the {basic} representation, the argument that is shared between the matrix predicate and the embedded predicate
is only attached to the matrix predicate. Thus, similarly as in the case of shared dependents in conjoined phrases,
there is no explicit relation between the embedded predicate and its subject. In the {enhanced} representation, this implicit subject relation is made explicit
with an additional relation, such as the \textsc{nsubj} relation\footnote{The fact that this relation is between an embedded predicate
and an argument of the matrix verb can be optionally marked with the \textsc{nsubj:xsubj} subtype.} below the sentence in (\ref{control}).

\enumsentence{
\raisebox{-8mm}{
\begin{dependency}
\begin{deptext}
Mary \& wants \& to \& buy \& a \& book \\
PROPN \& VERB \& PART \& VERB \& DET \& NOUN \\
\end{deptext}
\depedge[edge style=thick]{2}{1}{nsubj}
\depedge[edge style=thick]{4}{3}{mark}
\depedge[edge style=thick]{2}{4}{xcomp}
\depedge[edge style=thick]{6}{5}{det}
\depedge[edge style=thick]{4}{6}{obj}
\depedge[edge style=thick, edge below, edge unit distance=0.45em]{4}{1}{nsubj}
\end{dependency}
}
\label{control}
}

\paragraph{Relative Pronouns}

In the {enhanced} representation, the coreferential status of relative pronouns is marked with the special \textsc{ref} relation. Further,
to represent the implicit relation between the predicate of the relative clause and the antecedent of the relative pronoun, there is an additional
relation between the predicate and the antecedent, such as the \textsc{nsubj} relation between \textit{lived} and \textit{boy} in
(\ref{rel}).\footnote{The \textsc{nsubj} relation between \emph{lived} and \emph{who} is common to the basic and enhanced representation.}

\enumsentence{
\raisebox{-12mm}{
\begin{dependency}[column sep=0.4cm]
  \begin{deptext}
    the \& boy \& who \& lived \\
    DET \& NOUN \& PRON \& VERB \\
  \end{deptext}
     \depedge[edge style=thick]{2}{1}{det}
   \depedge[edge style=thick]{2}{4}{acl:relcl}
    \depedge[edge style=thick]{4}{3}{nsubj}
   \depedge[edge below, edge style=thick]{4}{2}{nsubj}
  \depedge[edge below, edge style=thick]{2}{3}{ref}
\end{dependency}
}
\label{rel}
}

\paragraph{Case Information}
Finally, since many modifier relation types such as \textsc{obl} or \textsc{acl} are used for many different types of relations, and since
adpositions or case information often disambiguate the semantic role, the {enhanced} representation provides augmented modifier relations that include
adposition or case information in the relation name, such as the \textsc{nmod:on} relation in (\ref{case}).

\enumsentence{
\raisebox{-2.5mm}{
\begin{dependency}[column sep=0.4cm]
  \begin{deptext}
    the \& house \& on \& the \& hill \\
    ADP \& NOUN \& ADP \& DET \& NOUN \\
  \end{deptext}
     \depedge{2}{1}{det}
     \depedge[edge style=thick]{5}{3}{case}
          \depedge[edge style=thick]{5}{4}{det}
   \depedge[edge style=thick]{2}{5}{nmod:on}
\end{dependency}
}
\label{case}
}

All enhancements are optional and users may decide to implement only a subset of these.
As of UD release v2.5, only 24 treebanks include an {enhanced} representation, and even fewer treebanks
implement all five enhancements (see also \newcite{droganova19}).
In many cases, the {enhanced} graphs can be computed automatically from a {basic} dependency tree (see \newcite{nivre18udw} for a discussion and evaluation of a rule-based and a machine learning-based converter from {basic} to {enhanced} dependencies), and \newcite{droganova19} recently used the Stanford Enhancer \cite{schuster16} to automatically predict {enhanced} dependencies for all UD treebanks.

\begin{figure*}
\centering
\includegraphics[width=0.95\textwidth]{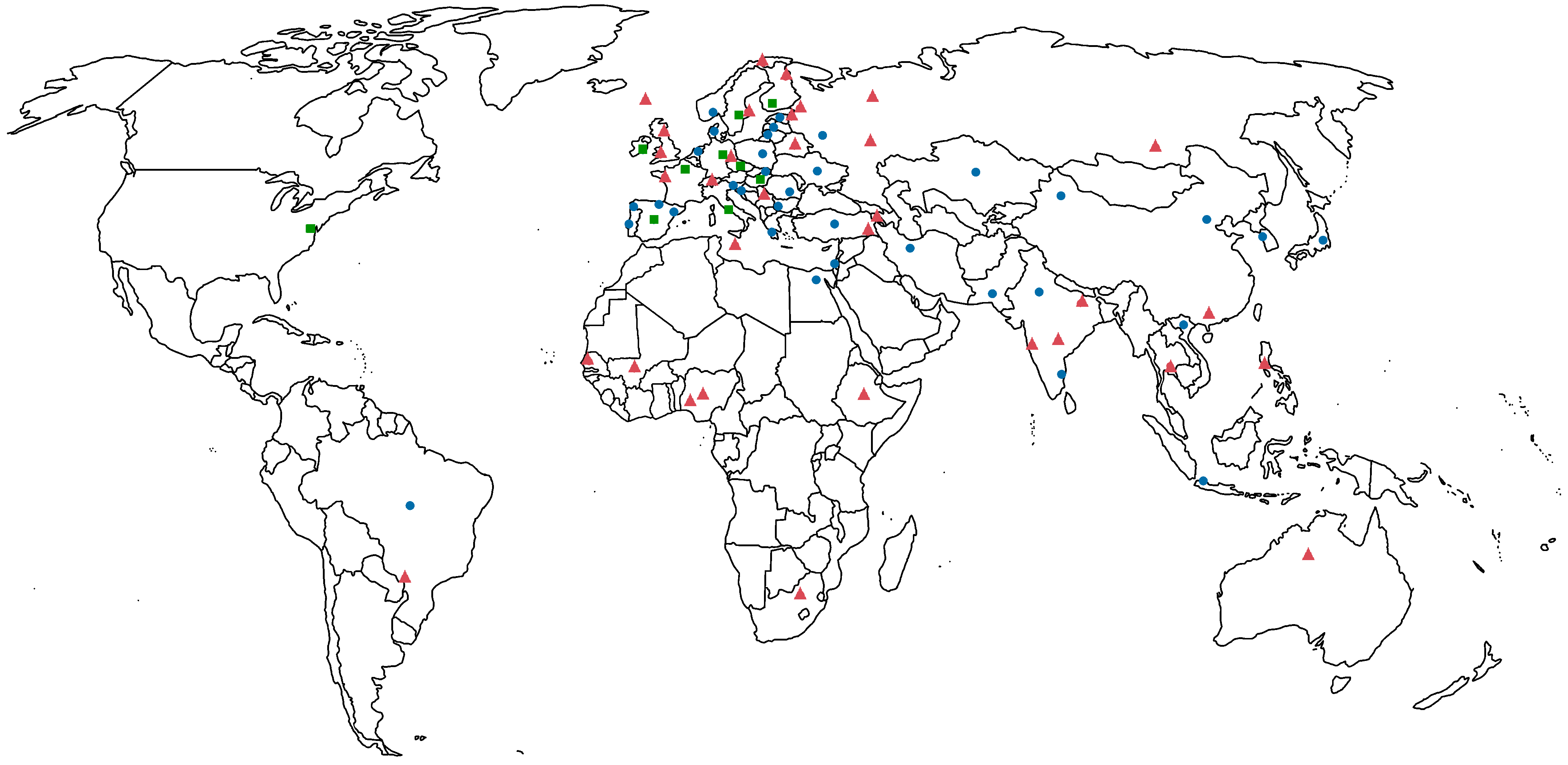}
\caption{Map of the world with language coverage of UD. Locations are approximate. Languages released in v1.0 of the
  collection (2015) are in \textcolor{MapGreen}{green {\small $\blacksquare$}}, those released in v2.0 (2017)
  are in \textcolor{MapBlue}{blue $\bigcdot$}~, and those released in v2.5 (2019) are in \textcolor{MapRed}{red $\blacktriangle$}. Coordinates are approximate based on the capital city or centre of the country where either the largest population of speakers lives, or where the treebank was created.}\label{fig:map}
\end{figure*}

\section{Available Treebanks}
UD release v2.5\footnote{UD releases are numbered by letting the first digit (2) refer to the version of the guidelines and the second digit (5) to the number of releases under that version.} \cite{zeman19ud25}
contains 157 treebanks representing 90 languages. Table~\ref{tab:treebanks} specifies for each language the number of treebanks available, as well as the total number of annotated sentences and words in that language. It is worth noting that the amount of data varies considerably between languages, from Skolt S\'ami with 36 sentences and 321 words, to German with over 200,000 sentences and nearly 4 million words. The majority of treebanks are small 
but it should be kept in mind that many of these treebanks are new initiatives and can be expected to grow substantially in the future.

The languages in UD v2.5 represent 20 different language families (or equivalent), listed in Table~\ref{tab:families}. The selection is very heavily biased towards Indo-European languages (48 out of 90), and towards a few branches of this family -- Germanic (10), Romance (8) and Slavic (13) -- but it is worth noting that the bias is (slowly) becoming less extreme over time.\footnote{The proportion of Indo-European languages has gone from 60\% in v2.1 to 53\% in v2.5.} Another way of visualizing the gradual extension of UD to new language families and geographic areas can be found in Figure~\ref{fig:map}, which shows the approximate geographic locations of languages added in UD v1.0 (green), UD v2.0 (blue) and UD v2.5 (red). It is clear that, whereas UD v1.0 was almost completely restricted to Europe, later versions have extended to other areas, and by v2.5 all inhabited continents are represented -- although there are still large white areas on the map.

The treebanks in UD v2.5 are also heterogeneous with respect to the type of text (or spoken data) annotated. A very coarse-grained picture of this variation can be gathered from Table~\ref{tab:genres}, which specifies the number of treebanks that contain some amount of data from different ``genres'', as reported by each treebank provider in the treebank documentation. The categories in this classification are neither mutually exclusive nor based on homogeneous criteria, but it is currently the best documentation that can be obtained.

\begin{table}[tb]
    \centering
\begin{footnotesize}
    \begin{tabular}{|llrr|}
    \hline
    \textbf{Family} &
    & & \textbf{Languages}\\
    \hline
    Afro-Asiatic & & & 7\\
    Austro-Asiatic & & & 1 \\
    Austronesian & & & 2 \\
    Basque & & & 1 \\
    Dravidian & & & 2\\
    Indo-European & & & 48 \\
    Japanese & & & 1 \\
    Korean & & & 1 \\
    Mande & & & 1 \\
    Mongolic & & & 1 \\
    Niger-Congo & & & 2\\
    Pama-Nyungan & & & 1 \\
    Sino-Tibetan & & & 3 \\
    Tai-Kadai & & & 1 \\
    Tupian & & & 1 \\
    Turkic & & & 3 \\
    Uralic & & & 11 \\
    \hline
    Code-Switching & & & 1 \\
    Creole & & & 1 \\
    Sign Language & & & 1 \\
   \hline
    \end{tabular}
\end{footnotesize}
    \caption{Language families
    in UD v2.5.}
    \label{tab:families}
\end{table}

\begin{table}[t]
    \centering
\begin{footnotesize}
    \begin{tabular}{|lrclr|}
    \hline
    \textbf{Genre} &\textbf{\#} & &  \textbf{Genre} & \textbf{\#} \\
    \hline
Academic &4  &~& News &98 \\
Bible &10  &~& Non-fiction &57 \\
Blog &17  &~& Poetry &4 \\
Email &2  &~& Reviews &7 \\
Fiction &42  &~& Social &9 \\
Grammar examples &13  &~& Spoken &18 \\
Learner essays &2  &~& Web &9 \\
Legal &22  &~& Wiki &46 \\
Medical &6  &~& & \\
\hline
    \end{tabular}
\end{footnotesize}
    \caption{Genres in UD v2.5 with number of treebanks.}
    \label{tab:genres}
\end{table}

\section{Conclusion}
The UD project has come a long way in only five years, and UD treebanks are now widely used in NLP as well as in linguistic research, especially with a typological orientation. Future priorities for the project include obtaining data from more languages -- in order to achieve better coverage of major language families -- but also obtaining more annotated data for existing languages -- in order to make the data more useful for NLP as well as linguistic studies. Finally, the work on achieving cross-linguistic consistency needs to continue. Adopting a common set of categories and guidelines is a first step in this direction, but ensuring that these are applied consistently across a growing set of typologically diverse languages will continue to be a challenge for years to come. Fortunately, efforts in this direction are constantly being pursued in the active UD user community.

\section{Acknowledgments}

We want to thank our colleagues in the UD core guidelines group Yoav Goldberg, Ryan McDonald, Slav Petrov and Reut Tsarfaty for fruitful discussions and comments on a draft version of this paper, 
as well as all the 345 UD treebank contributors, listed in \newcite{zeman19ud25}, without whom UD literally would not exist.

\section{Bibliographical References}
\bibliographystyle{lrec}
\bibliography{expanded,main}


\end{document}